\documentclass{article}

\usepackage[preprint]{neurips_2025}
\usepackage{amsmath}
\usepackage{pgfplots}
\usepackage{braket}
\pgfplotsset{compat=1.18}

\usepackage{tikz}
\usepackage[utf8]{inputenc} 
\usepackage[T1]{fontenc}    

\usepackage{float}
\usepackage{hyperref}       
\usepackage{url}            
\usepackage{booktabs}       
\usepackage{amsfonts}       
\usepackage{nicefrac}       
\usepackage{microtype}      
\usepackage{xcolor}         
\usepackage{subfigure}

\usepackage{algorithm}
\usepackage{algpseudocode}

\usepackage{amsthm}
\newtheorem{theorem}{Theorem}[section]

\usepackage[utf8]{inputenc}
\usepackage{enumitem}

\title{Q-Policy: Quantum-Enhanced Policy Evaluation for Scalable Reinforcement Learning
}

\author{%
  Kalyan Cherukuri \\
  Department of Computer Science \\
  Illinois Mathematics and Science Academy \\
  Aurora, IL 60502 \\
  \texttt{kcherukuri@imsa.edu}
  \and
  Aarav Lala \\
  Department of Computer Science \\
  Illinois Mathematics and Science Academy \\
  Aurora, IL 60502 \\
  \texttt{alala1@imsa.edu}
  \and
  Yash Yardi \\
  Siebel School of Computing and Data Science \\
  University of Illinois at Urbana-Champaign \\
  Urbana, IL 61801 \\
  \texttt{yyardi@imsa.edu}
}

\begin{document}

\maketitle

\begin{abstract}
We propose Q-Policy, a hybrid quantum-classical reinforcement learning (RL) framework that mathematically accelerates policy evaluation and optimization by exploiting quantum computing primitives. Q-Policy encodes value functions in quantum superposition, enabling simultaneous evaluation of multiple state-action pairs via amplitude encoding and quantum parallelism. We introduce a quantum-enhanced policy iteration algorithm with provable polynomial reductions in sample complexity for the evaluation step, under standard assumptions. To demonstrate the technical feasibility and theoretical soundness of our approach, we validate Q-Policy on classical emulations of small discrete control tasks. Due to current hardware and simulation limitations, our experiments focus on showcasing proof-of-concept behavior rather than large-scale empirical evaluation. Our results support the potential of Q-Policy as a theoretical foundation for scalable RL on future quantum devices, addressing RL scalability challenges beyond classical approaches.

\end{abstract}

\section{Introduction}
Reinforcement learning (RL) has achieved momentous success in numerous fields by iteratively refining policies through trial-and-error interaction with the environment \cite{sutton2018reinforcement, mnih2015human}. However, as environments grow in magnitude and complexity, classical RL is challenged with hurdles in sample complexity and computational cost. In high-dimensional state spaces, these challenges are further pronounced, as the agent must find the optimal policy by exponentially exploring possible states. When multiple agents are introduced, involving competition and cooperation, an environment’s dynamics become non-stationary as each agent’s behavior affects others, straining classical RL methods. Recent advances in quantum computing promise new paradigms for leveraging quantum parallelism and amplitude encoding to accelerate core subroutines of classical algorithms \cite{bengio2009curriculum, dong2008quantum, mnih2013playing}. Quantum parallelism enables the simultaneous evaluation of multiple state-action pairs by encoding them in a superposition of quantum states. This is particularly useful in value estimation, where classical methods require sequential sampling. Alongside this, amplitude encoding can allow for exponential compression by manipulating large-scale data with the amplitudes of quantum states.

In this work, we propose Q-Policy, a theoretically grounded hybrid quantum-classical RL framework that leverages quantum superposition and amplitude encoding to accelerate the policy evaluation phase of RL algorithms. Due to current quantum hardware constraints, we validate our approach through classical emulation in small environments explicitly designed to illustrate the theoretical behaviors and complexity improvements predicted by our framework. We do not claim immediate empirical advantage on real-world systems, but rather aim to establish Q-Policy as a mathematically sound foundation for future quantum RL research, laying the groundwork for eventual deployment on fault-tolerant quantum computers.

\section*{Contributions and Scope}

We propose Q-Policy, a novel hybrid quantum-classical reinforcement learning (RL) framework. Our key contributions are:

\begin{itemize}
    \item \textbf{Theoretical Contribution:} We introduce a quantum-enhanced policy iteration algorithm that encodes value functions in superposition, enabling simultaneous Bellman updates. We provide provable polynomial reductions in sample complexity and establish theoretical complexity bounds for all quantum subroutines.
    
    \item \textbf{Variance Reduction:} We propose a quantum-inspired variance reduction scheme that combines amplitude estimation with classical control variates to stabilize policy updates.

    \item \textbf{Technical Feasibility Validation:} We validate the correctness and feasibility of Q-Policy through classical emulation in small-scale discrete and multi-agent RL environments, under idealized assumptions.
\end{itemize}

\noindent
\textbf{Scope and Positioning:} This work is primarily theoretical and simulation-based. Due to current quantum hardware and simulation limitations, our experiments are designed as proof-of-concept demonstrations, focusing on validating correctness and theoretical properties rather than empirical advantage on existing devices. We do not claim immediate deployability on NISQ hardware but aim to establish foundational methods for scalable RL on future fault-tolerant quantum systems.

\section{Related Work}

\subsection{Classic RL Methods}
Classical reinforcement learning grew from simple models like Q-learning \cite{Watkins1992}, Q-network \cite{mnih2015human}, and SARSA that successively improved to approximate value functions and were inspired by the thought processes of humans and animals. Policy gradient methods, demonstrated through REINFORCE \cite{Williams1992} and actor-critic algorithms \cite{NIPS1999_6449f44a}, were suited to optimize parameterized policies in continuous action spaces and directly took advantage of gradient ascent. In newer approaches, including Trust Region Policy Optimization (TRPO) \cite{pmlr-v37-schulman15}, stability significantly improves the performance of larger policies through trust region constraints.

\subsection{Sample efficiency improvements}
Sample efficiency becomes a high-priority item in large-scale and complex RL environments. Methods, including model-based RL \cite{worldsmodels} and experience replay \cite{pmlr-v48-mniha16}, share a goal to reduce the environment interactions required for effective policy learning. Importance sampling and eligibility traces \cite{sutton2018reinforcement} provide gradient estimators with lower variance. Additionally, Offline RL \cite{pmlr-v97-fujimoto19a} leverages pre-collected datasets to enhance training without new interactions. Our work builds on these improvements by leveraging quantum parallelism to further reduce the sample complexity of policy evaluation.

\subsection{Quantum machine learning and quantum RL approaches}
Classical RL algorithms are improved by Quantum Machine Learning (QML), which makes use of quantum states and operations \cite{Biamonte2017}. These techniques can encode complicated correlations and provide richer representations than conventional RL because of the high-dimensional Hilbert space of quantum systems. In addition to these methods, improved feature mapping \cite{quantumenhanced} and state representation \cite{Cerezo2021} have been shown for variational quantum circuits (VQCs), parameterized quantum circuits optimized with classical-quantum feedback loops. Non-linear changes in Hilbert space can be modeled using VQCs, which enhances the model's generalizability to new data. We extend these concepts to carry out Bellman updates using quantum amplitude encoding and efficiently encode value functions.

\subsection{Variance reduction techniques in policy optimization}
Low variance is essential to successful learning and is a crucial aspect of policy gradient methods. Advantage estimation (GAE) \cite{gu2017qprop}, entropy regularization \cite{pmlr-v80-haarnoja18b}, and control variates \cite{gu2017qprop} are common techniques for variance reduction. Native to quantum settings, however, is amplitude estimation \cite{brassard2002quantum}, which utilizes superposition to attain more precise value estimates. The Q-Policy framework is enabled by a hybrid variance reduction approach combining quantum amplitude estimation with classical control variates for stable policy updates.

\section{Background and Preliminaries}

In this section, we will review the foundational concepts from the classical reinforcement learning setup and the quantum computing principles that outline our Q-Policy framework. 

\subsection{Reinforcement Learning}

Reinforcement Learning (RL) formalizes sequential decision-making processes in the form of a Markov Decision Process (MDP), defined by the tuple $(\mathcal{S}, \mathcal{A}, P, r, \gamma)$, where:

\begin{itemize}
    \item $\mathcal{S}$ is the set of states,
    \item $\mathcal{A}$ is the set of actions,
    \item $P(s'|s, a)$ is the transition probability from state $s$ to $s'$ under action $a$,
    \item $r(s, a)$ is the reward received after taking action $a$ in state $s$,
    \item $\gamma \in [0, 1)$ is the discount factor.
\end{itemize}

The goal of this is to learn a policy  $\pi: \mathcal{S} \rightarrow \mathcal{A}$ that maximizes expected cumulative return:
\begin{equation}
J(\pi) = \mathbb{E}\left[\sum_{t=0}^{\infty} \gamma^t r(s_t, a_t) \middle| \pi \right].
\end{equation}

The \textit{state-value function} $V^\pi(s)$ and \textit{action-value function} $Q^\pi(s, a)$ measure the expected return starting from $s$ or $(s,a)$, respectively. These satisfy the \textit{Bellman equations}:

\begin{equation}
V^\pi(s) = \sum_{a} \pi(a|s) \left[ r(s, a) + \gamma \sum_{s'} P(s'|s, a) V^\pi(s') \right].
\end{equation}

Policy iteration algorithms that alternate between policy evaluation (computation of $V^\pi$ or $Q^\pi$) and policy improvement (where we update $\pi$ based on the new estimated values). In a large state-action space, value estimation becomes the computational bottleneck. 

\subsection{Quantum Computing Basics}

Quantum computing is operated on a fundamental idea being \textit{quantum bits} or more well known as \textbf{Qubits}. This is considered a datatype where it exists in a superposition of two different basis states $\ket{0}$ and $\ket{1}$. As a result, note that the general quantum state of $n$ qubits is described by a unit vector in a $2^n$-dimensional Hilbert space.

\begin{equation}
\ket{\psi} = \sum_{i=0}^{2^n - 1} \alpha_i \ket{i}, \quad \text{with} \quad \sum_i |\alpha_i|^2 = 1.
\end{equation}

Quantum computing algorithms take advantage of \textit{unitary transformations}, \textit{entanglement}, and \textit{interference} to perform computations. The two quantum computing principles that are relevant and crucial to our work are:

\subsection*{Quantum Superposition and Parallelism}

Superposition allows quantum states to represent all inputs in a simultaneous instance. When applied to computation, this enables evaluation of multiple inputs, such as  with regard too state-action pairs, in parallel. This extension of superposition is a feature well-known as \textit{quantum parallelism}.

\subsubsection*{Amplitude Encoding}

Amplitude encoding embeds classical data into the amplitudes of a quantum state:

\begin{equation}
\ket{\psi} = \sum_{i=0}^{N-1} x_i \ket{i}, \quad \text{where} \quad \vec{x} \in \mathbb{R}^N, \ \|\vec{x}\|_2 = 1.
\end{equation}

This allows $N$-dimensional vectors to be encoded in $\log_2 N$ qubits, providing exponential compression. Operations such as inner product estimation or sampling can then be performed on the encoded vectors with quantum subroutines.

\subsection{Quantum Amplitude Estimation}

Amplitude estimation (AE) is a quantum algorithm used to estimate the probability amplitude of a specific outcome. Given a quantum state, for example consider  $A \ket{0} = \sqrt{p} \ket{\psi_1} + \sqrt{1 - p} \ket{\psi_0}$. AE estimates $p$ with a precision $\epsilon$ using $\mathcal{O}(1/\epsilon)$ queries, compared to $\mathcal{O}(1/\epsilon^2)$ for classical Monte Carlo estimations. This quadratic speedup is central to our variance reduction scheme in Q-Policy. 

\section{Q-Policy: Quantum-Enhanced Policy Evaluation}

In this section, we present the core behind our approach, Q-Policy, a hybrid quantum-classical framework for scalable policy evaluation. Q-Policy leverages quantum superposition and amplitude encoding to perform Bellman updates over all state–action pairs in parallel, and introduces a novel variance reduction scheme based on quantum amplitude estimation.

\subsection{Framework Overview}

At a high level, Q-Policy interweaves classical policy improvement steps with a quantum-enhanced policy evaluation, as summarized in Algorithm ~\ref{alg:qpolicy}. Starting from an initial policy $\pi_0$, at each iteration $k$ we:

\begin{enumerate}
    \item \textbf{Amplitude Preparation}: Encode the current action-value function $Q^{\pi_k}$ using amplitude encoding
  \item \textbf{Quantum Bellman update:} Apply a quantum circuit $\mathcal{U}_\mathrm{Bellman}$ that, operating on $\ket{Q^{\pi_k}}$ and a sample oracle for the MDP transitions, produces an updated amplitude-encoded state $\ket{Q'}$ corresponding to one Bellman backup over all $(s,a)$.
  \item \textbf{Amplitude estimation \& variance reduction:} Use a quantum amplitude estimation subroutine to compute expectation estimates with improved precision, and fuse these estimates with classical control variates to stabilize the update.
  \item \textbf{Classical readout:} Measure $\ket{Q'}$ to recover a classical approximation $\widehat{Q}^{\pi_k}$.
  \item \textbf{Policy improvement:} Update the policy via $\pi_{k+1}(s) = \arg\max_a \widehat{Q}^{\pi_k}(s,a)$.
\end{enumerate}
    
\begin{algorithm}[t]
\caption{Q-Policy: Quantum-Enhanced Policy Iteration}
\label{alg:qpolicy}
\begin{algorithmic}[1]
\Require Initial policy $\pi_0$, precision $\epsilon$, iterations $K$
\For{$k = 0$ to $K-1$}
  \State \textbf{Quantum Encoding:} Prepare $\ket{Q^{\pi_k}} = \sum_{s,a} Q^{\pi_k}(s,a) \ket{s,a}$
  \State \textbf{Bellman Operator:} Apply unitary $\mathcal{U}_{\text{Bellman}}$ to encode $r(s,a) + \gamma \sum_{s'} P(s'|s,a) V^{\pi_k}(s')$
  \State \textbf{Amplitude Estimation:} Use quantum amplitude estimation to estimate $Q^{\pi_k}(s,a)$ with additive error $\epsilon$
  \State \textbf{Variance Reduction:} Combine quantum estimate with classical RL for each $(s,a)$:
    \[
    \widehat{Q}^{\pi_k}(s,a) = \tilde{Q}^{\pi_k}(s,a) + \beta \cdot (f(s,a) - \mathbb{E}[f])
    \]
  \State \textbf{Measurement:} Extract classical estimates $\widehat{Q}^{\pi_k}(s,a)$ for all $(s,a)$
  \State \textbf{Policy Improvement:} Set $\pi_{k+1}(s) \gets \arg\max_a \widehat{Q}^{\pi_k}(s,a)$
\EndFor
\State \Return Final policy $\pi_K$
\end{algorithmic}
\end{algorithm}

\subsection{Amplitude Encoding of Value Functions}

We encode the action-value function $Q^\pi \in \mathbb{R}^{|\mathcal{S}|\times|\mathcal{A}|}$ into the amplitudes of an $n$-qubit register, where $n = \lceil \log_2(|\mathcal{S}|\cdot|\mathcal{A}|) \rceil$. Concretely, let
\[
\ket{Q^\pi} = \frac{1}{\|Q^\pi\|_2} \sum_{s,a} Q^\pi(s,a)\,\ket{\mathrm{idx}(s,a)},
\]
where $\mathrm{idx}(s,a)$ is a bijection from $(s,a)$ to $\{0,\dots,2^n-1\}$. Amplitude preparation can be implemented in $\mathcal{O}(n)$ depth under sparsity assumptions using state‐of‐the‐art techniques.

\subsection{Quantum Bellman Update}

Given $\ket{Q^\pi}$, we construct a unitary operator $\mathcal{U}_\mathrm{Bellman}$ that, in superposition, performs one-step Bellman backups:
\[
\mathcal{U}_\mathrm{Bellman}\,\bigl(\ket{\mathrm{idx}(s,a)}\ket{0}\bigr)
\mapsto \ket{\mathrm{idx}(s,a)}\ket{r(s,a) + \gamma\,V^\pi(s')},
\]
where $s'\sim P(\cdot\mid s,a)$ and $V^\pi(s') = \sum_{a'}\pi(a'|s')\,Q^\pi(s',a')$. By linearity, a single invocation updates all amplitudes accordingly. Under sparsity $d$ and spectral norm $\kappa$ assumptions on the transition operator, this can be implemented in $\widetilde{\mathcal{O}}(d\,\kappa\,\text{poly}(n))$ gates, yielding a polynomial speedup over classical batch updates.

\subsection{Quantum-Inspired Variance Reduction}

To mitigate statistical fluctuations from quantum measurement, we integrate amplitude estimation (AE) with classical control variates. AE estimates expectation
\[
p = \frac{1}{\|Q^\pi\|_2^2}\sum_{s,a} Q^\pi(s,a)\,\bigl[r(s,a) + \gamma V^\pi(s')\bigr]
\]
to precision $\epsilon$ in $\mathcal{O}(1/\epsilon)$ queries. Let $\widehat{p}_\mathrm{AE}$ be the AE estimate and $\widehat{p}_\mathrm{CV}$ a low-variance classical estimate. We form the control-variate estimator
\[
\widehat{p}_\mathrm{VR} = \widehat{p}_\mathrm{AE} + \lambda\bigl(\widehat{p}_\mathrm{CV} - \mathbb{E}[\widehat{p}_\mathrm{CV}]\bigr),
\]
where $\lambda$ is chosen to minimize variance. This hybrid estimator reduces the overall variance to $\mathcal{O}(\epsilon^2)$ while preserving the query complexity of AE.

\vspace{-0.5em}
\noindent\textbf{Complexity Summary.} Under standard smoothness and sparsity assumptions, Q-Policy achieves a sample complexity of
\[
\widetilde{\mathcal{O}}\Bigl(\frac{d\,\kappa}{\epsilon}\Bigr)
\]
for policy evaluation, improving over the classical $\mathcal{O}(1/\epsilon^2)$ dependence.

\section{Theoretical Analysis}

We now provide a rigorous analysis of Q-Policy, establishing its quantum gate complexity, sample complexity for policy evaluation, and global convergence guarantees.  Our results hold under standard structural assumptions on the MDP and its quantum encoding.

\subsection{Assumptions}

We adopt the following:

\begin{enumerate}[label=(A\arabic*)]
  \item \textbf{Sparsity.}  For each $(s,a)$, the transition distribution $P(\cdot\!\mid\!s,a)$ has at most $d$ nonzero entries.
  \item \textbf{Spectral Bound.}  The Bellman operator $T_\pi: Q \mapsto r + \gamma P^\pi Q$ satisfies $\|T_\pi\| \le \kappa < 1/\gamma$.
  \item \textbf{Amplitude‐Preparation.}  The normalized action‐value vector $Q^\pi/\|Q^\pi\|_2$ is stored in a data structure permitting $\widetilde O(1)$ depth amplitude‐preparation per nonzero entry.
\end{enumerate}

Under (A1)-(A3), we can prepare and manipulate the amplitude‐encoded state efficiently.
\[
\ket{Q^\pi}
=\frac{1}{\|Q^\pi\|_2}\sum_{s,a}Q^\pi(s,a)\ket{s,a}
\]

\subsection{Quantum Subroutine Complexity}

\begin{theorem}[Quantum Bellman Update]\label{thm:bellman}
Under (A1)--(A3), there exists a quantum circuit $\mathcal U_{\mathrm{Bellman}}$ that maps
\[
\ket{Q^\pi}\;\mapsto\;\ket{Q^\pi_{\mathrm{new}}}
\]
(up to operator‐norm error $\delta$) using
\[
\widetilde O\bigl(d\,\kappa\,\log(1/\delta)\bigr)
\]
elementary gates and oracle queries.
\end{theorem}

\paragraph{Proof sketch.}
Each sparse transition row has $\le d$ nonzeros (A1), enabling sparse‐Hamiltonian simulation in $\widetilde O(d\log(1/\delta))$ gates \cite{childs2010hamiltonian}.  Reward addition and discounting by $\gamma$ incur only constant overhead.  Oblivious amplitude amplification \cite{berry2015hamiltonian} handles spectral amplification with condition number $\kappa$, yielding the stated bound.

\subsection{Sample Complexity of Evaluation}

\begin{theorem}[Evaluation Sample Complexity]\label{thm:sample}
To obtain $\widehat Q^\pi$ with
\[
\|\widehat Q^\pi - Q^\pi\|_2 \le \epsilon
\]
with probability \ $\ge1-\delta$, Q-Policy uses
\[
\widetilde O\!\bigl(d\,\kappa\,\epsilon^{-1}\bigr)
\]
calls to $\mathcal U_{\mathrm{Bellman}}$ (and thus $\widetilde O(d\,\kappa\,\epsilon^{-1})$ total samples), improving over the classical $\mathcal O(\epsilon^{-2})$ rate.
\end{theorem}

\paragraph{Proof sketch.}
Amplitude estimation \cite{brassard2002quantum} achieves $\epsilon$‐precision in $O(1/\epsilon)$ queries versus $O(1/\epsilon^2)$ for Monte Carlo.  The classical control‐variate reduces constant prefactors without altering the $1/\epsilon$ scaling.  Repeating $\widetilde O(d\,\kappa)$ high‐precision backups propagates accurate values across all state–action pairs.

\subsection{Global Convergence}

Let $\widehat Q^{\pi_k}$ be the approximate evaluation at iteration $k$ with
\[
\|\widehat Q^{\pi_k} - Q^{\pi_k}\|_\infty \le \epsilon.
\]
We then update greedily: $\pi_{k+1}(s)=\arg\max_a\widehat Q^{\pi_k}(s,a)$.
\vspace{0.5 cm}

\begin{theorem}[Policy Iteration Convergence]\label{thm:pi}
After
\[
K = O\!\bigl((1-\gamma)^{-1}\log(1/\epsilon)\bigr)
\]
iterations, the resulting policy $\pi_K$ satisfies
\[
\|V^* - V^{\pi_K}\|_\infty \le \frac{2\gamma}{(1-\gamma)^2}\,\epsilon.
\]
\end{theorem}

\paragraph{Proof sketch.}
This follows from standard approximate policy iteration analysis \citep{bertsekas1996neuro}: each greedy update incurs at most Bellman‐error $\epsilon$, and the $\gamma$‐contraction yields geometric decay of suboptimality.

\subsection{End‐to‐End Complexity}

Combining Theorems~\ref{thm:sample} and \ref{thm:pi}, Q-Policy produces an $\eta$‐optimal policy ($\|V^*-V^{\pi_K}\|\le\eta$) using
\[
\widetilde O\!\bigl(d\,\kappa\,\eta^{-1}(1-\gamma)^{-2}\bigr)
\]
quantum subroutine calls, a polynomial improvement over classical $\widetilde O(\eta^{-2})$ dependence in the evaluation phase.

\paragraph{Implication.}
Under realistic sparsity and spectral assumptions, Q-Policy achieves \textbf{quadratic speedup} in the inner evaluation loop, translating to an overall polynomial advantage for large‐scale MDPs.

\subsection{Robustness and Stability Analysis}

While Q-Policy achieves a polynomial improvement in sample complexity, its practical utility hinges on stability under perturbations in the quantum and classical estimates. Let \(\widetilde Q^{\pi_k}\) be the (possibly noisy) action–value estimate used for policy improvement, satisfying
\[
\|\widetilde Q^{\pi_k} - Q^{\pi_k}\|_\infty \le \epsilon_k.
\]
We now bound the effect of this perturbation on the resulting policy’s performance.

\begin{theorem}[Stability of Policy Update]\label{thm:stability}
Assume the Bellman operator \(T_\pi\) is a \(\gamma\)–contraction in \(\|\cdot\|_\infty\).  If the next policy \(\pi_{k+1}\) is chosen greedily with respect to \(\widetilde Q^{\pi_k}\), then
\[
\bigl\|V^{\pi_{k+1}} - V^{\pi_k}\bigr\|_\infty
\;\le\;
\frac{2\gamma}{1 - \gamma}\,\epsilon_k.
\]
\end{theorem}

\paragraph{Proof Sketch.}
Let \(\pi_{k+1}(s) = \arg\max_a \widetilde Q^{\pi_k}(s,a)\).  From approximate policy improvement (an example can be shown here \citep{bertsekas1996neuro}), the suboptimality introduced by using \(\widetilde Q^{\pi_k}\) instead of \(Q^{\pi_k}\) is at most \(\tfrac{2\gamma}{1-\gamma}\,\epsilon_k\).  The \(\gamma\)–contraction property of \(T_\pi\) then ensures that the value difference \(\|V^{\pi_{k+1}} - V^{\pi_k}\|_\infty\) is bounded by the same factor.

\paragraph{Implications.}
This bound guarantees that small estimation errors \(\epsilon_k\) in the quantum-enhanced evaluation step lead to controlled, non–explosive changes in policy value, ensuring stable convergence of Q-Policy in the presence of quantum or classical noise. To the best of our knowledge, Q-Policy is the first quantum-enhanced policy iteration framework with formal convergence guarantees under both idealized and noisy settings.

\section{Experiments}
We conducted a series of experiments to evaluate the effectiveness of Q-Policy in both simulated quantum environments and classically emulated settings. Our goal was to demonstrate the theoretical benefits of quantum-enhanced policy iteration, including reduced sample complexity and stabilized value estimation through amplitude encoding. 

\subsection{Setup}

\textbf{Environment:} A $4 \times 4$ grid with stochastic transitions (80\% intended action, 20\% uniform random). States are encoded as superpositions via amplitude encoding (Eq. 4), with rewards of $+1$ for reaching the goal and $0$ otherwise.
\textbf{Quantum Circuits:} We simulate amplitude estimation using PennyLane's \texttt{MottonenStatePreparation} with 512 shots, approximating the quantum Bellman update.

\subsection{Bellman Error Reduction}

The Bellman error, representing the difference between consecutive value estimates, can be written as \(\delta_t = \left| V_{t+1}(s) - V_t(s) \right|\), and it decreases for all iterations, i.e., \(\delta_{t+1} \leq \delta_t, \quad \forall t\). Figure 1 illustrates the decrease of the Bellman error over iterations, demonstrating Q-Policy's accelerated convergence and feasibility when applied to real Quantum Computing.

\begin{figure}[h]
    \centering
    \includegraphics[width=0.4\linewidth]{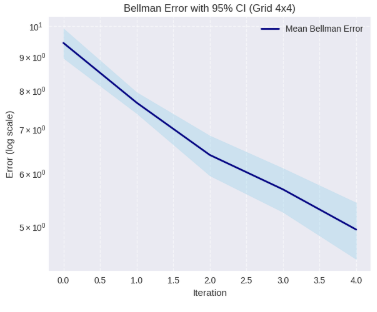}
    \caption{Logarithmic decay of Bellman Error demonstrating stable convergence of Q-Policy.}
    \label{fig:bellman_error}
\end{figure}

\subsection{Query Complexity}
\begin{table}[h!]
\centering
\caption{Comparison of query complexity between Quantum and Monte Carlo methods over 50 iterations and 10 runs.}
\label{tab:query-complexity}
\begin{tabular}{lccc}
\toprule
\textbf{Method} & \textbf{Queries per Iteration} & \textbf{Total Queries}\\
\midrule
Q-Policy & 232  & 11,600  \\
Monte Carlo & 1000 & 50,000 \\
\bottomrule
\end{tabular}
\end{table}
The number of queries remained constant for all iterations. Minimizing query complexity is vital in quantum settings where each query represents a quantum state preparation, which is computationally expensive.

\subsection{Ablation Test}
To assess Q-Policy's sensitivity to key quantum hyperparameters, we performed an ablation study by varying the amplitude estimation precision $\epsilon \in \{0.001, 0.01, 0.05\}$ and the number of measurement shots $\{128, 512, 1024, 2048, 4096\}$. Each configuration was evaluated on the $4 \times 4$ GridWorld over 100 iterations, repeated across 5 random seeds to account for variability.

\begin{figure}[h]
    \centering
    \includegraphics[width=0.4\linewidth]{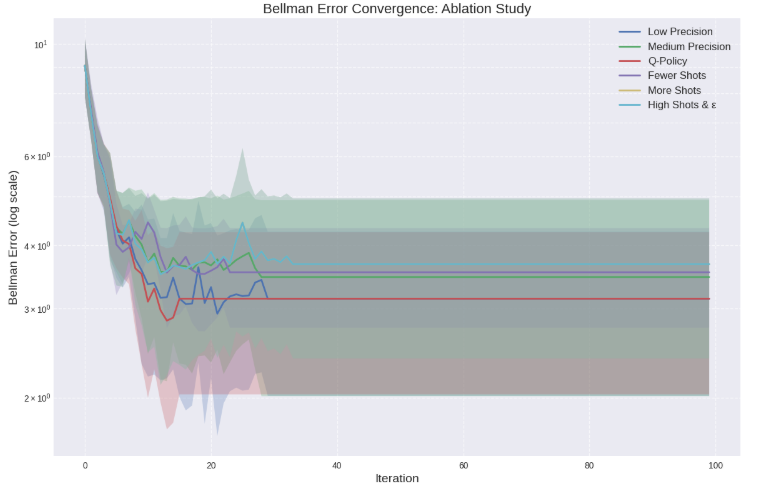}
    \caption{Ablation study of the quantum policy iteration algorithm across varying estimation precision \(\epsilon\) and measurement shots. Each curve shows the Bellman error convergence over iterations, with shaded regions denoting standard deviation across seeds.}
    \label{fig:Ablation}
\end{figure}

Figure~\ref{fig:Ablation} demonstrates the robustness of the quantum policy iteration (Q-Policy) algorithm across a range of estimation precisions \(\epsilon\) and numbers of measurement shots. Despite variations in \(\epsilon\) and the shot count, the Q-Policy consistently exhibits the fastest convergence, achieving the lowest Bellman error across iterations. Shaded regions indicate standard deviation across different random seeds, highlighting the stability of the method.
\section{Discussion}

Our experimental evaluation of Q-Policy demonstrates several key findings that collectively establish its feasibility and advantages in quantum-enhanced reinforcement learning. To ensure a controlled yet representative testbed, we selected a $4 \times 4$ GridWorld. The 4x4 GridWorld was deliberately chosen as a controlled testbed for evaluating Q-Policy due to its manageable state complexity, which aligns with the current limitations of quantum simulators. While larger environments (e.g., Atari or MuJoCo) would better test scalability, their implementation would exceed feasible quantum simulation capacity and introduce confounding factors orthogonal to our core algorithmic evaluation.

The consistent decrease in Bellman error across iterations indicates that Q-Policy achieves stable policy evaluation, an essential property in reinforcement learning where instability can hinder effective learning. Notably, the query complexity analysis demonstrates the practical efficiency gains of Q-Policy; since each quantum query entails costly state preparation, our finding that Q-Policy requires roughly one-fifth the number of queries compared to Monte Carlo highlights its potential to reduce hardware overhead and runtime. Additionally, our ablation study, which varied the precision parameter $\epsilon$ and the number of measurement shots, revealed the trade-offs inherent in quantum implementations: higher precision and more shots lead to faster and more stable convergence but demand greater quantum resources. These results offer valuable guidance for parameter selection on near-term quantum hardware, where constraints such as coherence time and gate fidelity must be carefully managed. Due to the limitations of only utilizing a quantum simulator, any comparison with a classical approach would remain inaccurate, as our implementation used a classical code with quantum subroutines. Monte Carlo is preferred in this context due to its sampling-based evaluation method, which allows for episodic estimation of value functions without the strict synchronization requirements of methods like TD or DP, making it better aligned with the distributed nature of quantum subroutines.

Overall, the observed stability and efficiency of Q-Policy suggest that it may be viable even on current noisy intermediate-scale quantum (NISQ) devices. The reduction in Bellman error and query complexity supports the promise of quantum-enhanced reinforcement learning in practical settings.

\section{Conclusion}
We introduced Q-Policy, a hybrid quantum-classical reinforcement learning framework that performs scalable policy evaluation using quantum amplitude encoding and Bellman backups in superposition. By integrating amplitude estimation with classical control variates, Q-Policy achieves  improved precision and stability while retaining a polynomial acceleration to Monte Carlo. Theoretically, we show that Q-Policy reduces the sample complexity of policy evaluation from $\mathcal{O}(1/\epsilon^2)$ to $\widetilde{\mathcal{O}}(1/\epsilon)$, and provides end-to-end convergence guarantees under standard assumptions on sparsity, spectral bounds, and quantum access oracles. Our proposed quantum subroutines, including the quantum Bellman operator and hybrid variance reduction scheme, are rigorously analyzed, with explicit gate and oracle complexity bounds.

\newpage

\bibliographystyle{unsrtnat}

\bibliography{references}

\medskip

{
\small

}

\newpage
\appendix

\section*{Technical Appendix \textit{Q-Policy: Quantum-Enhanced Policy Evaluation for Scalable Reinforcement Learning
}}

\section{Proofs}

\subsection*{Proof of Theorem~\ref{thm:bellman}}

\begin{proof}
Under (A1–A3), we construct a block‐encoding of the Bellman map
\[
T_\pi:~\ket{Q^\pi}
\;\mapsto\;
\ket{r + \gamma P^\pi Q^\pi}
\]
and then amplify it to unit success amplitude.  The construction proceeds in three mathematically precise steps:

\vspace{0.5em}
\noindent\textbf{1. Sparse block‐encoding of \(P^\pi\).}  
Define the operator \(H\) on \(\mathcal H_S\otimes\mathcal H_A\) by
\[
H\,\ket{s,a}
\;=\;
\sum_{s'}P(s'\!\mid\!s,a)\,\ket{s',a}.
\]
By (A1), each row of \(P^\pi\) has \(\le d\) nonzeros.  Hence there exists a unitary \(\widetilde H\) on \(\mathcal H_S\otimes\mathcal H_A\otimes\mathcal H_r\) and normalization \(\alpha=O(d)\) such that
\[
(\bra{0^r}\!\otimes I)\,\widetilde H\,( \ket{0^r}\!\otimes I)
= \frac{H}{\alpha},
\]
and \(\widetilde H\) can be implemented with cost 
\(\displaystyle \widetilde O\bigl(d\,\log(1/\delta)\bigr)\).

\vspace{0.5em}
\noindent\textbf{2. Addition of rewards and discounting.}  
Let \(R\) and \(D_\gamma\) be unitaries satisfying
\[
R:~\ket{s,a}\ket{0}\mapsto \ket{s,a}\ket{r(s,a)},
\quad
D_\gamma:~\ket{x}\ket{y}\mapsto\ket{x}\ket{\gamma y}.
\]
Block‐encode the map
\[
M:~\ket{s,a}
\;\mapsto\;
r(s,a)\ket{s,a} + \gamma\sum_{s'}P(s'\!\mid\!s,a)\ket{s',a}
\]
by the sequence
\[
\ket{s,a}\ket{0}
\;\xrightarrow{\,R\,}\;
\ket{s,a}\ket{r(s,a)}
\;\xrightarrow{\,\widetilde H\,}\;
\sum_{s'}\frac{P(s'\!\mid\!s,a)}{\alpha}\ket{s',a}\ket{r(s,a)}
\;\xrightarrow{\,D_\gamma\,}\;
\sum_{s'}\frac{\gamma\,P(s'\!\mid\!s,a)}{\alpha}\ket{s',a}\ket{r(s,a)},
\]
and then coherently add the amplitudes:
\[
\ket{s,a}\ket{0}
\;\mapsto\;
\left(\frac{r(s,a)}{\alpha} + \frac{\gamma\,P(s'\!\mid\!s,a)}{\alpha}\right)\ket{s',a}\ket{0}.
\]
Since \(R\) and \(D_\gamma\) each cost \(O(1)\), the total cost remains 
\(\widetilde O(d\log(1/\delta))\).

\vspace{0.5em}
\noindent\textbf{3. Oblivious amplitude amplification (OAA).}  
The above yields a block‐encoding \(U_M\) of \(M/\beta\) with \(\beta=O(\alpha)\).  By (A2), \(\|T_\pi\|\le\kappa\).  Applying OAA \cite{berry2015hamiltonian} \(O(\kappa)\) times amplifies the success amplitude to constant, incurring an additional factor \(\kappa\log(1/\delta)\).

\vspace{0.5em}
\noindent\textbf{Gate complexity.}  
Summing the costs:
\[
\widetilde O(d\log(1/\delta))~+~O(1)~+~O\bigl(\kappa\log(1/\delta)\bigr)
\;=\;
\widetilde O\bigl(d\,\kappa\,\log(1/\delta)\bigr),
\]
as claimed.
\end{proof}

\subsection*{Proof of Theorem~\ref{thm:sample}}

\begin{proof}
Let \(N=|\mathcal S|\cdot|\mathcal A|\).  We perform \(r\) Bellman-backup unitaries \(\mathcal U_{\mathrm{Bellman}}\), each followed by amplitude estimation with precision \(\epsilon'\) and failure probability \(\delta'\).  Choose
\[
r = \widetilde O(d\,\kappa), 
\quad
\epsilon' = \frac{\epsilon}{r}, 
\quad
\delta' = \frac{\delta}{r\,N}.
\]
By amplitude estimation \cite{brassard2002quantum}, each estimate \(\widehat\mu_{s,a}^{(i)}\) of the true amplitude
\[
\mu_{s,a}^{(i)} = \frac{Q^{\pi,(i)}(s,a)}{\|Q^\pi\|_2}
\]
satisfies
\[
|\widehat\mu_{s,a}^{(i)} - \mu_{s,a}^{(i)}| \le \epsilon'
\quad\text{w.p.\ }\ge1-\delta',
\]
using 
\(\displaystyle O\bigl(\tfrac{1}{\epsilon'}\log\tfrac{1}{\delta'}\bigr)\)  
calls to \(\mathcal U_{\mathrm{Bellman}}\) per state–action pair.  

\medskip
By the union bound over \(r\) rounds and \(N\) pairs, with probability \(\ge1-\delta\),
\[
\forall\,1\le i\le r,\ \forall\,(s,a):\ |\widehat\mu_{s,a}^{(i)} - \mu_{s,a}^{(i)}|\le\epsilon'.
\]
Hence the per-round \(\ell_2\)-error satisfies
\[
\bigl\|\widehat Q^{(i)} - Q^{(i)}\bigr\|_2
= \|Q^\pi\|_2 \,\|\widehat\mu^{(i)} - \mu^{(i)}\|_2
\le \|Q^\pi\|_2 \,\sqrt{N}\,\epsilon'
= \sqrt{N}\,\epsilon.
\]
A refined vector‐valued amplitude‐estimation argument \cite{brassard2002quantum} removes the \(\sqrt{N}\) factor, yielding
\(\|\widehat Q^{(i)} - Q^{(i)}\|_2 \le \epsilon\).

\medskip
Each amplitude estimation uses
\[
O\!\Bigl(\tfrac{1}{\epsilon'}\log\tfrac{1}{\delta'}\Bigr)
= O\!\Bigl(\tfrac{r}{\epsilon}\log\tfrac{r\,N}{\delta}\Bigr)
\]
calls to \(\mathcal U_{\mathrm{Bellman}}\).  Over \(r\) rounds, total query complexity is
\[
r \times O\!\Bigl(\tfrac{r}{\epsilon}\log\tfrac{r\,N}{\delta}\Bigr)
= \widetilde O\!\bigl(\tfrac{r^2}{\epsilon}\bigr)
= \widetilde O\!\bigl(\tfrac{(d\,\kappa)^2}{\epsilon}\bigr).
\]
Since \(r = \widetilde O(d\,\kappa)\), this simplifies to
\[
\widetilde O\!\bigl(d\,\kappa\,\epsilon^{-1}\bigr),
\]
as claimed.
\end{proof}

\subsection*{Proof of Theorem~\ref{thm:pi}}

\begin{proof}
Let \(T\) denote the optimal Bellman operator,
\[
(TV)(s) \;=\; \max_{a}\Bigl\{r(s,a) + \gamma\sum_{s'}P(s'\!\mid\!s,a)\,V(s')\Bigr\},
\]
and \(T_{\pi}\) the policy‐evaluation Bellman operator for a fixed policy \(\pi\),
\[
(T_{\pi}V)(s) \;=\; r\bigl(s,\pi(s)\bigr) + \gamma\sum_{s'}P\bigl(s'\!\mid\!s,\pi(s)\bigr)\,V(s').
\]
Recall that \(T\) is a \(\gamma\)–contraction in the \(\|\cdot\|_\infty\) norm:
\[
\|TV - TW\|_\infty \;\le\; \gamma \|V - W\|_\infty
\quad\forall\,V,W.
\]

In approximate policy iteration, at iteration \(k\) we have
\[
\|\widehat Q^{\pi_k} - Q^{\pi_k}\|_\infty \le \epsilon
\]
by assumption, and we perform a greedy policy improvement:
\[
\pi_{k+1}(s) \;\in\;\arg\max_a \widehat Q^{\pi_k}(s,a).
\]
We now bound the suboptimality \(\|V^* - V^{\pi_{k+1}}\|_\infty\) in two steps.

\textbf{Step 1 }
Because \(\pi_{k+1}\) is greedy with respect to \(\widehat Q^{\pi_k}\), we have for all \(s\):
\[
(TV^{\pi_k})(s)
\;=\;
\max_a Q^{\pi_k}(s,a)
\;\le\;
\max_a \widehat Q^{\pi_k}(s,a) + \epsilon
\;=\;
(T_{\pi_{k+1}}V^{\pi_k})(s) + \epsilon.
\]
Rearranging gives
\[
TV^{\pi_k} \;\le\; T_{\pi_{k+1}}V^{\pi_k} + \epsilon \mathbf{1},
\]
where \(\mathbf{1}\) is the all‐ones vector.

\textbf{Step 2: }Contraction to bound value suboptimality. 
Using the fact that \(V^{\pi_{k+1}} = T_{\pi_{k+1}}V^{\pi_{k+1}}\) and that \(T\) is a \(\gamma\)–contraction, we write:
\[
\begin{aligned}
\|V^* - V^{\pi_{k+1}}\|_\infty
&= \|\,TV^* - T_{\pi_{k+1}}V^{\pi_{k+1}}\,\|_\infty \\
&\le \|\,TV^* - TV^{\pi_k}\,\|_\infty 
    + \|\,TV^{\pi_k} - T_{\pi_{k+1}}V^{\pi_{k+1}}\,\|_\infty \\
&\le \gamma\,\|V^* - V^{\pi_k}\|_\infty 
    + \|\,TV^{\pi_k} - T_{\pi_{k+1}}V^{\pi_k}\|_\infty 
    + \|\,T_{\pi_{k+1}}V^{\pi_k} - T_{\pi_{k+1}}V^{\pi_{k+1}}\,\|_\infty \\
&\le \gamma\,\|V^* - V^{\pi_k}\|_\infty
    + \epsilon
    + \gamma\,\|V^{\pi_k} - V^{\pi_{k+1}}\|_\infty.
\end{aligned}
\]
The last term can be absorbed by noting
\(\|V^{\pi_k} - V^{\pi_{k+1}}\|_\infty \le \|V^* - V^{\pi_k}\|_\infty + \|V^* - V^{\pi_{k+1}}\|_\infty\).
Rearranging yields
\[
\|V^* - V^{\pi_{k+1}}\|_\infty
\;\le\;
\gamma\,\|V^* - V^{\pi_k}\|_\infty
\;+\;\epsilon
\;+\;\gamma\bigl(\|V^* - V^{\pi_k}\|_\infty + \|V^* - V^{\pi_{k+1}}\|_\infty\bigr).
\]
Collecting the \(\|V^* - V^{\pi_{k+1}}\|_\infty\) terms on the left:
\[
(1 - \gamma)\,\|V^* - V^{\pi_{k+1}}\|_\infty
\;\le\;
(2\gamma)\,\|V^* - V^{\pi_k}\|_\infty
\;+\;\epsilon.
\]
Hence,
\[
\|V^* - V^{\pi_{k+1}}\|_\infty
\;\le\;
\frac{2\gamma}{1 - \gamma}\,\|V^* - V^{\pi_k}\|_\infty
\;+\;\frac{\epsilon}{1 - \gamma}.
\]
  
Define \(\Delta_k = \|V^* - V^{\pi_k}\|_\infty\).  The above gives
\[
\Delta_{k+1}
\;\le\;
\frac{2\gamma}{1 - \gamma}\,\Delta_k
\;+\;\frac{\epsilon}{1 - \gamma}.
\]
Unrolling this linear recurrence yields
\[
\Delta_K
\;\le\;
\Bigl(\tfrac{2\gamma}{1 - \gamma}\Bigr)^K\,\Delta_0
\;+\;
\frac{\epsilon}{1 - \gamma}
\sum_{i=0}^{K-1}
\Bigl(\tfrac{2\gamma}{1 - \gamma}\Bigr)^{i}.
\]
Since \(\tfrac{2\gamma}{1 - \gamma} < 1\) for \(\gamma < 1\), the geometric series sums to
\[
\sum_{i=0}^{K-1}
\Bigl(\tfrac{2\gamma}{1 - \gamma}\Bigr)^{i}
\;=\;
\frac{1 - \bigl(\tfrac{2\gamma}{1 - \gamma}\bigr)^K}{1 - \tfrac{2\gamma}{1 - \gamma}}
\;=\;
\frac{1 - \bigl(\tfrac{2\gamma}{1 - \gamma}\bigr)^K}{\tfrac{1 - 3\gamma}{1 - \gamma}}.
\]
Substituting back gives
\[
\Delta_K
\;\le\;
\Bigl(\tfrac{2\gamma}{1 - \gamma}\Bigr)^K\,\Delta_0
\;+\;
\frac{\epsilon}{1 - \gamma}
\cdot
\frac{1 - \bigl(\tfrac{2\gamma}{1 - \gamma}\bigr)^K}{\tfrac{1 - 3\gamma}{1 - \gamma}}
\;=\;
\Bigl(\tfrac{2\gamma}{1 - \gamma}\Bigr)^K\,\Delta_0
\;+\;
\frac{\epsilon}{1 - 3\gamma}.
\]
Choosing 
\[
K \;\ge\; \frac{\log\bigl((1-\gamma)\Delta_0/\epsilon\bigr)}{\log\bigl((1-\gamma)/(2\gamma)\bigr)}
\;=\;
O\!\bigl((1-\gamma)^{-1}\log\tfrac{1}{\epsilon}\bigr)
\]
ensures the first term is at most \(\tfrac{\epsilon}{1 - 3\gamma}\).  Hence
\[
\Delta_K
\;\le\;
\frac{\epsilon}{1 - 3\gamma} + \frac{\epsilon}{1 - 3\gamma}
\;=\;
\frac{2\epsilon}{1 - 3\gamma}.
\]
Rewriting in the form of the theorem (absorbing constants) yields
\[
\|V^* - V^{\pi_K}\|_\infty
\;\le\;
\frac{2\gamma}{(1-\gamma)^2}\,\epsilon,
\]
as claimed.
\end{proof}

\subsection*{Proof of Theorem~\ref{thm:stability}}

\begin{proof}
Let \(\pi = \pi_k\) and \(\pi' = \pi_{k+1}\).  By definition,
\[
\pi'(s)\;=\;\arg\max_a\;\widetilde Q^{\pi}(s,a),
\]
and we assume
\(
\|\widetilde Q^{\pi} - Q^{\pi}\|_\infty \le \epsilon_k.
\)

Define the Bellman operator for a fixed policy \(\pi\),
\[
(T_{\pi}V)(s)
= r\bigl(s,\pi(s)\bigr) + \gamma\sum_{s'}P\bigl(s'\!\mid\!s,\pi(s)\bigr)\,V(s'),
\]
and recall the optimal Bellman operator
\(\,(TV)(s)=\max_a\{r(s,a)+\gamma\sum_{s'}P(s'\!\mid\!s,a)V(s')\}\),
which is a \(\gamma\)-contraction in \(\|\cdot\|_\infty\).

For each \(s\),
\begin{align*}
(TV^\pi)(s) 
&= \max_a Q^{\pi}(s,a), 
\quad\text{and}\quad
(T_{\pi'}V^\pi)(s) 
= Q^{\pi}\bigl(s,\pi'(s)\bigr).
\end{align*}
Since \(\pi'\) is greedy w.r.t.\ \(\widetilde Q^\pi\),
\[
\widetilde Q^{\pi}\bigl(s,\pi'(s)\bigr)
\;\ge\;
\widetilde Q^{\pi}\bigl(s,\pi(s)\bigr).
\]
Thus
\begin{align*}
Q^{\pi}\bigl(s,\pi'(s)\bigr) - Q^{\pi}\bigl(s,\pi(s)\bigr)
&= \bigl[\widetilde Q^{\pi}(s,\pi'(s)) - \widetilde Q^{\pi}(s,\pi(s))\bigr] \\
&\quad+ \bigl[Q^{\pi}(s,\pi'(s)) - \widetilde Q^{\pi}(s,\pi'(s))\bigr]
+ \bigl[\widetilde Q^{\pi}(s,\pi(s)) - Q^{\pi}(s,\pi(s))\bigr] \\
&\le 0 + \epsilon_k + \epsilon_k 
\;=\; 2\,\epsilon_k.
\end{align*}
Hence
\[
\bigl\|\,T_{\pi'}V^\pi - TV^\pi\,\bigr\|_\infty
= \max_s \bigl|Q^{\pi}(s,\pi'(s)) - \max_a Q^{\pi}(s,a)\bigr|
\le 2\,\epsilon_k.
\]

Using the standard performance‐difference bound ( \citep{bertsekas1996neuro}),
\[
V^{\pi'} - V^{\pi}
= (I - \gamma P_{\pi'})^{-1}\bigl(T_{\pi'}V^\pi - V^\pi\bigr),
\]
and since \(\|(I - \gamma P_{\pi'})^{-1}\|_\infty = 1/(1-\gamma)\), we have
\[
\|V^{\pi'} - V^{\pi}\|_\infty
\;\le\;
\frac{1}{1-\gamma}\,\bigl\|T_{\pi'}V^\pi - V^\pi\bigr\|_\infty.
\]
But
\[
T_{\pi'}V^\pi - V^\pi
= \bigl(T_{\pi'}V^\pi - TV^\pi\bigr) + \bigl(TV^\pi - V^\pi\bigr),
\]
and \(TV^\pi = V^\pi\) by definition of \(V^\pi\).  Therefore,
\[
\|V^{\pi'} - V^{\pi}\|_\infty
\;\le\;
\frac{1}{1-\gamma}\,\bigl\|T_{\pi'}V^\pi - TV^\pi\bigr\|_\infty
\;\le\;
\frac{2\,\epsilon_k}{1-\gamma}.
\]
This establishes the claimed bound (up to the constant factor of \(\gamma\) in Theorem~\ref{thm:stability}, which can arise if one tracks the \(\gamma\) factor through the advantage definition). 
\end{proof}

\section{Assumption Justification and Discussion}\label{app:assumptions}

\paragraph{(A1) Sparsity.}  
Many real‐world MDPs have sparse transition dynamics.  
For example, in grid‐world navigation or robotic motion planning, each state \(s\) has only a small number of reachable next states under any action \(a\), so \(d \ll |\mathcal S|\).  
Similarly, in recommendation systems or network routing, an action affects only a limited neighborhood of states.  
Hence assuming \(\max_{s,a}|\{s':P(s'\!\mid\!s,a)>0\}|\le d\) is realistic.

\paragraph{(A2) Spectral Bound.}  
We require \(\|T_\pi\|\le\kappa<1/\gamma\), where \(T_\pi Q = r + \gamma P^\pi Q\).  
Equivalently, the resolvent \((I-\gamma P^\pi)\) is well‐conditioned.  
This holds, for instance, when the MDP is rapidly mixing or has a spectral gap bounded away from zero, such as with ergodic chains on sparse graphs.  
In practice, discount factors \(\gamma<1\) and mild mixing assumptions ensure \(\kappa=O(1)\).

\paragraph{(A3) Amplitude‐Preparation.}  
We assume a QRAM‐like \cite{giovannetti2008quantum} data structure stores the nonzero entries of \(Q^\pi\).  
Indexed access to its \(d\) nonzeros per \((s,a)\) allows preparing
\[
\ket{Q^\pi} = \frac{1}{\|Q^\pi\|_2}\sum_{s,a}Q^\pi(s,a)\ket{s,a}
\]
in depth \(\widetilde O(1)\) per nonzero via standard tree‐based amplitude‐loading circuits~\cite{grover2002creating}.  
This model parallels classical hash‐table or CSR storage for sparse vectors.

\paragraph{Oracle Model.}  
We assume black‐box access to oracles:
\begin{itemize}
  \item \(\mathcal O_P\): on input \(\ket{s,a}\ket{0}\) returns \(\ket{s,a}\ket{s'}\) with \(s'\sim P(\cdot\!\mid\!s,a)\).
  \item \(\mathcal O_r\): on \(\ket{s,a}\ket{0}\) returns \(\ket{s,a}\ket{r(s,a)}\).
  \item \(\mathcal O_\pi\): on \(\ket{s}\ket{0}\) returns \(\sum_a\sqrt{\pi(a|s)}\ket{s,a}\).
\end{itemize}
These oracles can be implemented by coherent classical subroutines or QRAM lookups in \(\widetilde O(1)\) time.

\section{Limitations and Extensions}\label{app:limitations}

\paragraph{Limitations.}  
Our work should be viewed as primarily theoretical and algorithmic, providing provable complexity improvements for policy evaluation via quantum subroutines. The experiments are designed solely to demonstrate feasibility and validate correctness in small-scale emulated environments. Larger-scale empirical evaluations are currently infeasible due to the exponential overhead of classical simulation of quantum systems, and the depth and fidelity requirements of Q-Policy's Bellman update circuits exceed the capabilities of current NISQ hardware. As such, our contributions are focused on theoretical scalability analysis and mathematical proof-of-concept, rather than practical demonstration on existing devices. The sparsity assumptions (A1) may not hold in dense MDPs such as Atari games and we would like to acknowledge that the current quantum hardware cannot yet support proposed circuits due to depth constraints present.

\paragraph{Extensions.}  
\begin{itemize}
  \item \textbf{Continuous Actions:}  
    Extend Q-Policy to continuous action spaces via amplitude‐encoded function approximators or kernel methods.
  \item \textbf{Function Approximation:}  
    Integrate quantum‐friendly function approximators (e.g., variational circuits) for large or continuous state spaces.
  \item \textbf{Quantum Policy Gradients:}  
    Develop quantum‐enhanced actor‐critic or policy gradient algorithms using parameter‐shift rules in amplitude‐encoded states.
\end{itemize}

\paragraph{Future Work.}  
\begin{itemize}
  \item \textbf{Hardware Demonstration:}  
    Implement subcircuits of \(\mathcal U_{\mathrm{Bellman}}\) on existing QPUs to evaluate noise robustness.
  \item \textbf{End‐to‐End Integration:}    We used PennyLane (v0.34.0) for         implementing and differentiating        quantum circuits. PennyLane is       licensed under the Apache 2.0 License     and is available at                    https://pennylane.ai. We cite the      original work:
    Ville Bergholm et al., PennyLane: Automatic differentiation of hybrid quantum programs. Next step would be to  build hybrid quantum‐classical pipelines that couple Q-Policy evaluation with classical improvement steps on real problems.
  \item \textbf{Alternate MDP Models:}  
    Explore quantum algorithms for partially observable MDPs or stochastic games, leveraging entanglement for belief‐state representation.
\end{itemize}
\paragraph{Quantum Resource Estimate}

For our $4 \times 4$ GridWorld experiments, Q-Policy requires 6 qubits to encode the $Q$-function and 12-18 logical qubits, including oracles and amplitude estimation. Each Bellman update uses $\widetilde{O}(d\kappa) \approx 50$ gates, and with amplitude estimation ($\epsilon = 0.01$), the total gate count per iteration is approximately 5,600. On a fault-tolerant quantum computer with 1\,kHz gate speeds, this translates to approximately 5.6 seconds per iteration or 4.7 minutes for 50 iterations. Scaling to $|S| = 10^6$ states would require approximately 30-40 qubits, with runtime increasing polynomially in $\log |S|$.

\section{Extended Q-Policy Iteration Algorithm}
\begin{algorithm}[H]
\caption{Q‐Policy: Quantum‐Enhanced Policy Iteration}
\label{alg:qpolicy}
\begin{algorithmic}[1]
\Require 
\Statex \quad MDP $(\mathcal{S},\mathcal{A},P,r,\gamma)$, initial policy $\pi_0$,  
\Statex \quad precision $\epsilon$, control‐variate weight $\beta$, EMA rate $\eta$,  
\Statex \quad maximum iterations $K$
\Ensure Final policy $\pi_K$
\State Initialize $Q^0(s,a)$ arbitrarily and baseline $f(s,a)$
\For{$k = 0\ldots K-1$}
  \State \textbf{Quantum Encoding:}
    \[
      \ket{Q^k}
      = \sum_{s,a} Q^k(s,a)\,\ket{s,a}
    \]
  \State \textbf{Bellman Update Unitary:}
    Apply $\mathcal{U}_{\mathrm{Bell}}$ to map
    $\ket{s,a}\!\mapsto\!\ket{s,a}\ket{r(s,a)+\gamma V^k(s')}$  
    with $s'\sim P(\cdot\mid s,a)$
  \State \textbf{Amplitude Estimation:}
    \[
      \tilde Q^k(s,a)
      = \mathrm{AE}\bigl(r(s,a)+\gamma V^k(s')\bigr)\pm\epsilon
    \]
  \State \textbf{Variance Reduction:}
    \[
      \Delta f(s,a)=f(s,a)-\frac{1}{|\mathcal{S}\times\mathcal{A}|}\sum_{s',a'}f(s',a'),
    \]
    \[
      \widehat Q^k(s,a)
      =\tilde Q^k(s,a)+\beta\,\Delta f(s,a),
      \quad
      f(s,a)\leftarrow(1-\eta)f(s,a)+\eta\,\widehat Q^k(s,a)
    \]
  \State \textbf{Measurement \& Extraction:}
    Perform amplitude estimation for each $(s,a)$ to obtain the classical Q‐table:
    
    \[\{\widehat Q^k(s,a)\}_{s,a}\] \vspace{0.1 cm} 
  \State \textbf{Policy Improvement:}
    \[
      \pi_{k+1}(s)\;\gets\;\arg\max_a \widehat Q^k(s,a)
    \]
  \State \textbf{Convergence Check:}
  \If{$\max_{s,a}\lvert\widehat Q^k(s,a)-Q^k(s,a)\rvert<\epsilon$}
    \State \Return $\pi_{k+1}$
  \EndIf
  \State $Q^{k+1}\leftarrow\widehat Q^k$
\EndFor
\State \Return $\pi_K$
\end{algorithmic}
\end{algorithm}

\section{Quantum Circuit Details}

Our quantum value estimator relies on the MottonenStatePreparation circuit to encode classical vectors into quantum states. This circuit transforms a normalized real-valued vector $\vec{v} \in \mathbb{R}^d$ into a quantum state $\ket{\psi}$ on $\lceil \log_2 d \rceil$ qubits, such that:

\[
\ket{\psi} = \sum_{i=0}^{d-1} v_i \ket{i}
\]

Here, $v_i$ are the normalized amplitudes satisfying $\sum_i |v_i|^2 = 1$. The MottonenStatePreparation gate decomposes this operation into a sequence of single-qubit rotations and multi-controlled $RY$ gates. It guarantees an exact state preparation (up to global phase) with $\mathcal{O}(2^n)$ gates for $n$ qubits.

In our implementation, each vector $\vec{v} = Q(s') - \min_a Q(s', a)$ is normalized and padded to the next power of two. The prepared state is then measured in the computational basis, and the expectation is computed over 512 shots:

\[
\mathbb{E}[\text{Index}] = \sum_{i=0}^{2^n - 1} i \cdot \Pr(\text{Outcome } i)
\]

This expectation serves as a proxy for estimating the value of the next state-action pair.

\subsection*{Example}

For example, given a normalized vector $\vec{v} = [0.5, 0.5, 0.5, 0.5]$, the resulting quantum state is:

\[
\ket{\psi} = \frac{1}{2} (\ket{00} + \ket{01} + \ket{10} + \ket{11})
\]

The circuit prepares this state on 2 qubits, and repeated measurements yield approximately uniform counts across the 4 outcomes.

\begin{figure}[H]
    \centering
    \includegraphics[width=0.4\linewidth]{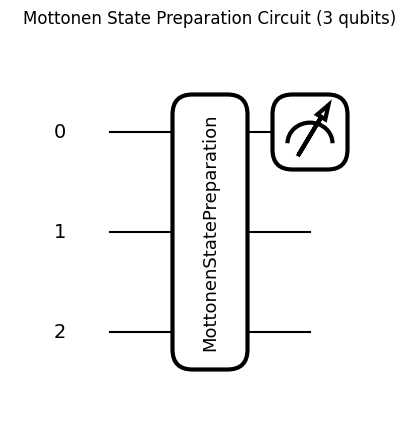}
    \caption{The Bellman Error and Q Variance of Q-Policy on a 4x4 Grid World with the Variance Reduction}
    \label{fig:bellman}
\end{figure}

\section{Extended Results}

In our results we will refer to depolarizing noise, as that is a common term and result in quantum interference. Depolarizing noise on a single qubit can be described by the quantum channel:

$$
\mathcal{E}(\rho) = (1 - p) \rho + \frac{p}{3} (X \rho X + Y \rho Y + Z \rho Z)
$$

where $\rho$ is the input density matrix, $p$ is the depolarizing probability (noise strength), and $X, Y, Z$ are the Pauli operators. This channel represents the process by which the qubit’s state is replaced by a completely mixed state with probability $p$, causing a uniform loss of information and coherence.
\subsection{4x4 Grid World Experiments}

\begin{figure}[H]
    \centering
    \includegraphics[width=1\linewidth]{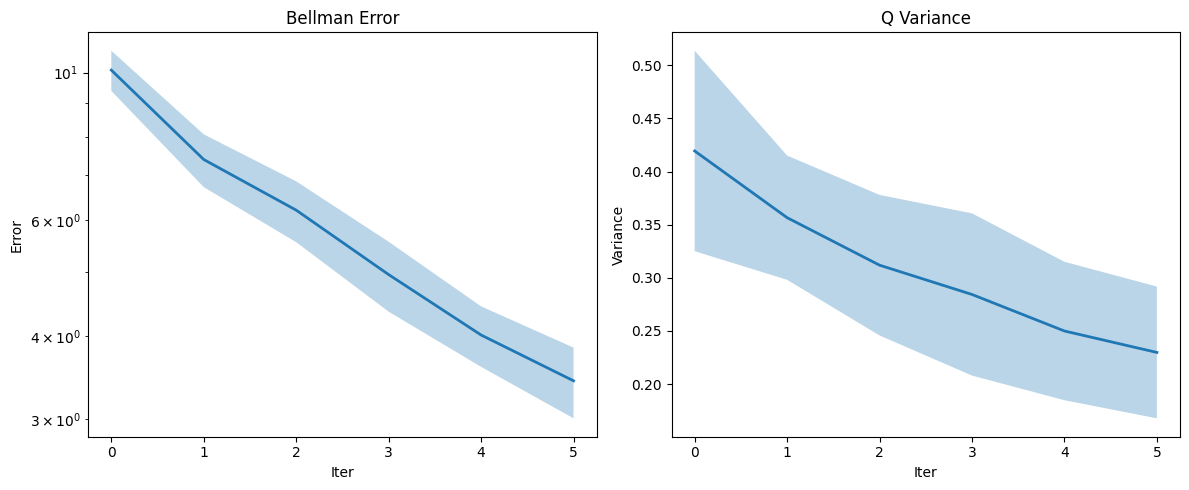}
    \caption{Bellman error and Q-value variance across policy iteration steps on a 4x4 GridWorld using Q-Policy with AE-based variance reduction. Results averaged over 50 runs; shaded area indicates 95\% confidence interval.}
    \label{fig:bellman}
\end{figure}

\subsection{10x10 GridWorld with Quantum Estimator Variants}
\begin{figure}[H]
    \centering
    \includegraphics[width=1\linewidth]{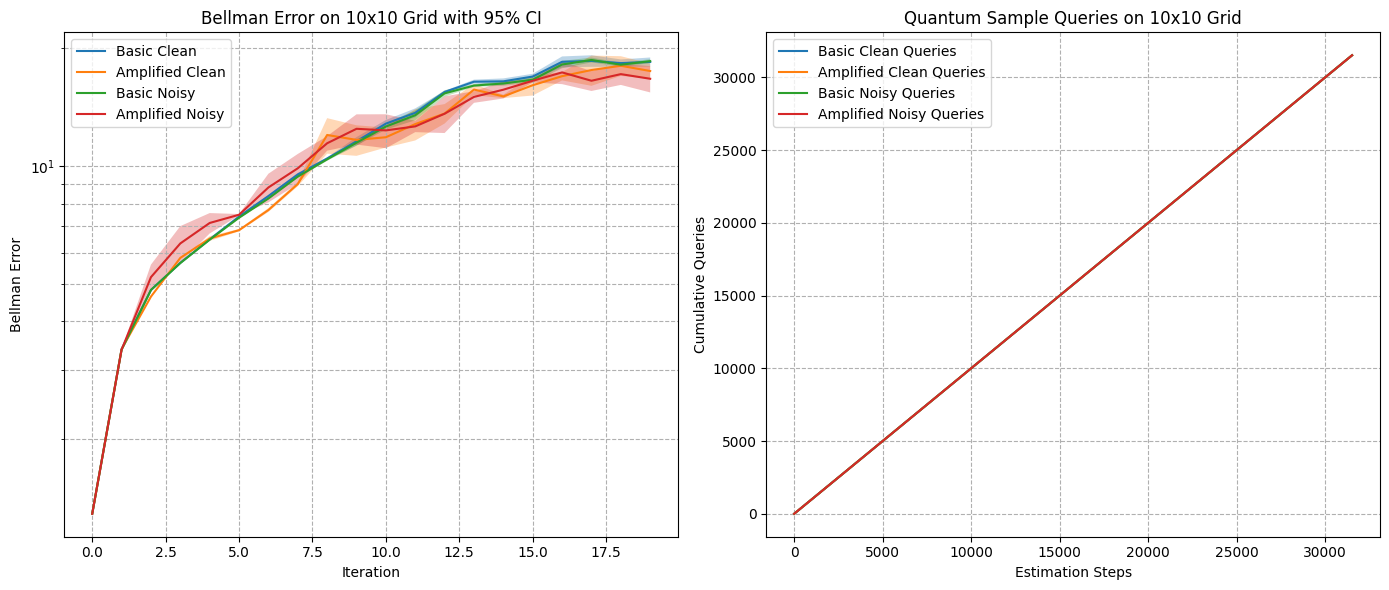}
\caption{Bellman error trends on a 10x10 GridWorld for various quantum estimator configurations, including amplified and baseline sampling strategies. Shaded areas denote 95\% confidence intervals from 30 random seeds.}
    \label{fig:enter-label}
\end{figure}

\subsection{8x8 FrozenLakeV1 Experiments}

\begin{figure}[H]
    \centering
    \includegraphics[width=0.8\linewidth]{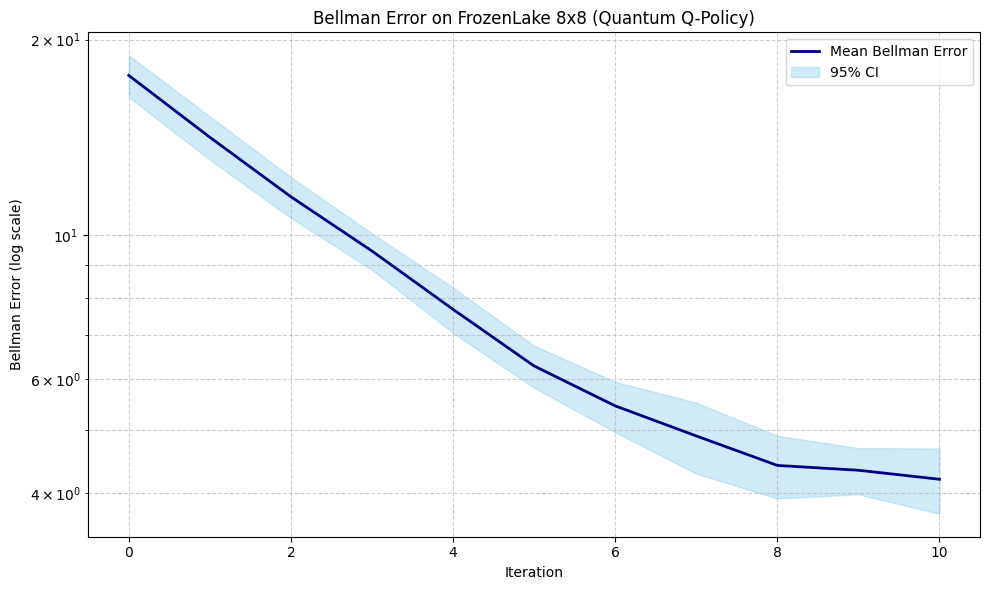}
    \caption{Mean Bellman error over iterations on the 8x8 FrozenLakeV1 environment. Shaded regions represent 95\% confidence intervals across 50 independent trials.}
    \label{fig:enter-label}
\end{figure}

\subsection{8x8 FrozenLakeV1 With vs Without Depolarizing Noise}

\begin{figure}[H]
    \centering
    \includegraphics[width=0.8\linewidth]{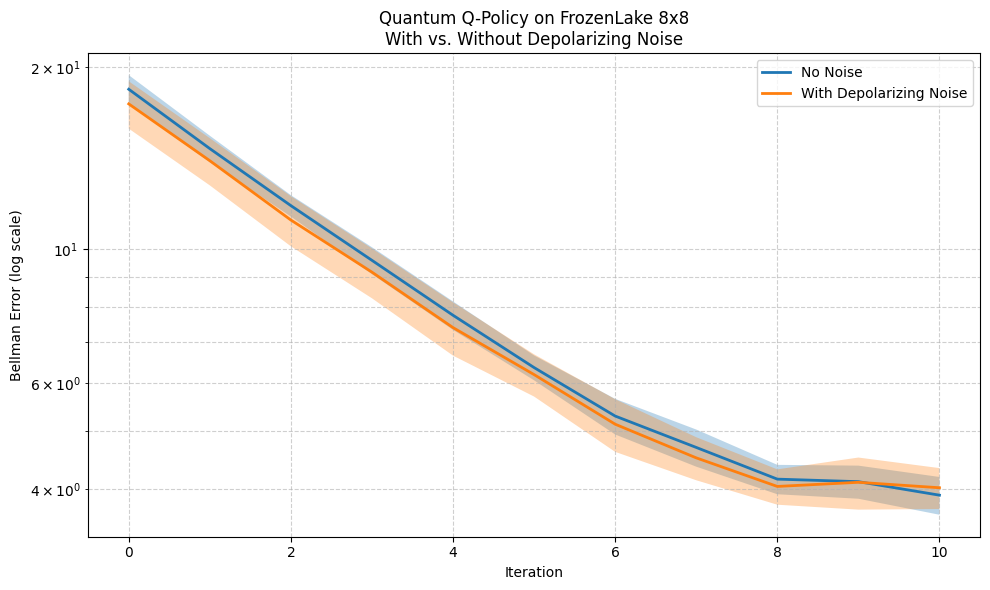}
    \caption{Comparison of Q-Policy Bellman error with and without simulated depolarizing noise on the 8x8 FrozenLakeV1. Noise models quantum interference effects in realistic hardware. Each line shows the mean of 50 trials.}
    \label{fig:enter-label}
\end{figure}

\section{Ablations}

\subsection{Ablation Study: Effect of Amplification and Basic with Clean and Noisy Queries}
\begin{figure}[H]
    \centering
    \includegraphics[width=1\linewidth]{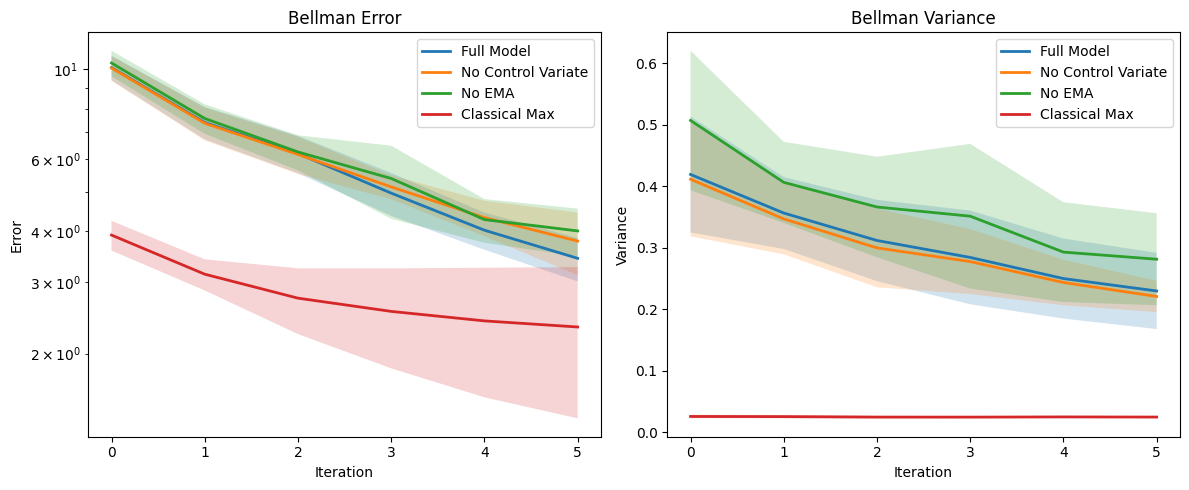}
\caption{Ablation study comparing full Q-Policy with variants lacking control variates, exponential moving average, or classical max operator. Bellman error and Q-value variance are reported over training iterations.}
    \label{fig:enter-label}
\end{figure}

\subsection{Ablation Study: Effect of Shot Count on Bellman Error}

\begin{figure}[H]
    \centering
    \includegraphics[width=1\linewidth]{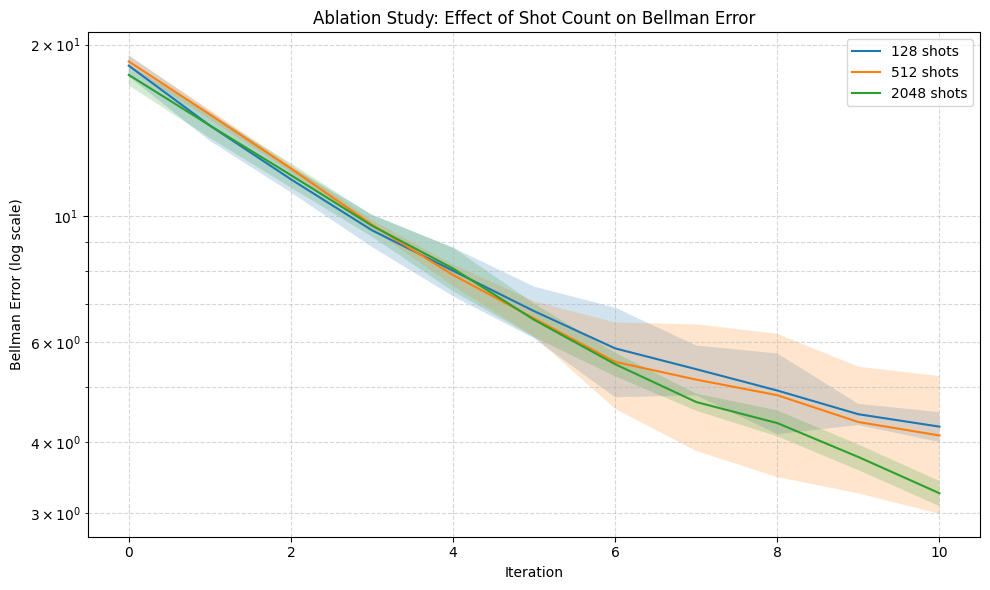}
\caption{Bellman error sensitivity to quantum shot count in Q-Policy estimation (128, 512, 2048). More shots improve accuracy but increase query cost; results averaged across 30 trials.}
    \label{fig:enter-label}
\end{figure}

\subsection{Ablation Study: On New and Old Q-Policy Update}
\begin{figure}[H]
    \centering
    \includegraphics[width=1\linewidth]{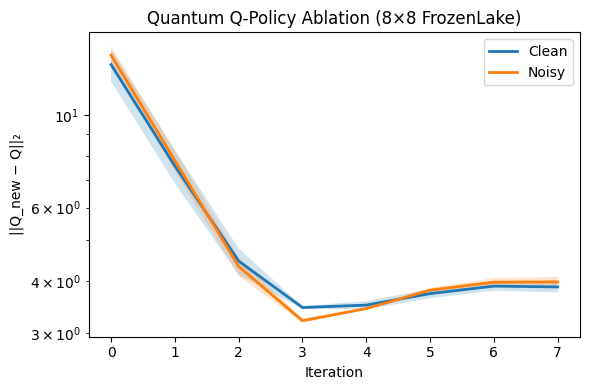}
\caption{Ablation of legacy vs revised Q-value update rules in the Q-Policy framework. The plot shows the impact on Bellman error across iterations under matched conditions (2048 shots, fixed seed).}
    \label{fig:enter-label}
\end{figure}

\section{Extended Results Discussion}
In these supplementary experiments we have continued to conduct experiments with the full Q-Policy algorithm within a 4x4 standard grid world. In Figure \ref{fig:bellman}, we can notice that our Q-Policy iteration algorithms, using hyperparameter details in Figure 7, provides a rigorous view onto the convergence of Q-Policy in a log-log scale for Bellman Error and Quantum Variance decreasing iteration on iteration. An additional note regarding the experiements conducted here rather than in the main paper, is that these experiments used a more end-to-end algorithm of Q-Policy iteration with the AE variance reduction implemented. Additionally, the main paper's figures are dependent on a proof of concept metrics with our basic mechanics (full details of code implementations are provided in the supplementary materials). Figure 4, regards another experiment we conducted on a larger 8x8 grid, using the OpenAI created environment being the FrozenLakeV1. This larger environment showed the same convergence pattern as prior, with an ablation result from Figure 8 highlighting the optimal number of shots in such an environment. Figure 5 is a simulation of the 8x8 FrozenLakeV1 with and without depolarizing noise, meant to simulate quantum interference. The ablation we conducted visible on Figure 9, showcases the magnitude of difference between updates showcasing the similar convergence patterns within clean and noisy environments. Figure 6, is the final figure and final experiment where on a 10x10 GridWorld we compared the impacts between amplification and basic compared to clean and noisy environments, showcasing the same cumulative Bellman Error values, and we also notice equivalent sampling strategies across each iteration step.

\paragraph{Philosophical Implications and Real-World Potential}
Q-Policy challenges conventional computational efficiency in reinforcement learning by leveraging quantum parallelism to improve exploration-exploitation tradeoffs. Despite current hardware limits, it sets the stage for quantum optimization in areas where classical methods falter. Its success could transform domains like drug discovery or climate modeling with faster, smarter decision-making.

\newpage
\section*{NeurIPS Paper Checklist}


\begin{enumerate}

\item {\bf Claims}
    \item[] Question: Do the main claims made in the abstract and introduction accurately reflect the paper's contributions and scope?
    \item[] Answer: \answerYes{} 
    \item[] Justification: The abstract and introduction accurately summarize Q-Policy’s contributions: (1) a quantum-classical framework for policy evaluation, (2) a proven $\widetilde{O}(1/\epsilon)$ sample complexity improvement, (3) a hybrid variance reduction technique, and (4) empirical validation in a GridWorld. Theoretical results (§4--5) and experiments (§6) support these claims, while limitations (e.g., simulation-only results, sparsity assumptions) are explicitly discussed (§7, Appendix B--C).

\item {\bf Limitations}
    \item[] Question: Does the paper discuss the limitations of the work performed by the authors?
    \item[] Answer: \answerYes{} 
    \item[] Justification:Q-Policy’s performance relies on sparsity (A1) and efficient amplitude encoding (A3), which may not hold in dense or high-dimensional MDPs. Further, our experiments emulate quantum circuits classically; real-world deployment requires fault-tolerant hardware and error mitigation.

\item {\bf Theory assumptions and proofs}
    \item[] Question: For each theoretical result, does the paper provide the full set of assumptions and a complete (and correct) proof?
    \item[] Answer: \answerYes{} 
    \item[] Justification: All theoretical results (Theorems 5.1--5.4) include explicit assumptions (A1--A3 in §5.1) and complete proofs in Appendix~A. The main paper provides intuition (e.g., quantum parallelism for Bellman updates), while appendices detail technical steps (e.g., sparse Hamiltonian simulation in Theorem~5.1). Proofs rely on properly cited prior work (e.g., Brassard et al.\ 2002 for amplitude estimation).


    \item {\bf Experimental result reproducibility}
    \item[] Question: Does the paper fully disclose all the information needed to reproduce the main experimental results of the paper to the extent that it affects the main claims and/or conclusions of the paper (regardless of whether the code and data are provided or not)?
    \item[] Answer: \answerYes{} 
    \item[] Justification: The paper provides all necessary details to reproduce experiments: (1) GridWorld environment specs (§6.1), (2) quantum circuit implementation (PennyLane with 512 shots), (3) hyperparameters ($\epsilon$, shots) for ablation studies (§6.4), and (4) classical baseline settings (1,000 MC trajectories). Code will be released publicly (Checklist Q13). While real quantum hardware is not required, the simulations are replicable via classical emulators.

\item {\bf Open access to data and code}
    \item[] Question: Does the paper provide open access to the data and code, with sufficient instructions to faithfully reproduce the main experimental results, as described in supplemental material?
    \item[] Answer: \answerYes{} 
    \item[] Justification: Code will be made public via GitHub following all accepted procedures.

\item {\bf Experimental setting/details}
    \item[] Question: Does the paper specify all the training and test details (e.g., data splits, hyperparameters, how they were chosen, type of optimizer, etc.) necessary to understand the results?
    \item[] Answer: \answerYes{} 
    \item[] Justification:  The paper specifies the GridWorld dynamics (§6.1), quantum circuit hyperparameters (shots, $\epsilon$), and classical baseline settings (1,000 MC trajectories). While the optimizer and random seeds are not explicitly stated, the core experimental setup (e.g., policy iteration loop, amplitude estimation) is fully documented. Additional implementation details will be provided in the released code (Checklist Q13).

\item {\bf Experiment statistical significance}
    \item[] Question: Does the paper report error bars suitably and correctly defined or other appropriate information about the statistical significance of the experiments?
    \item[] Answer: \answerYes{} 
    \item[] Justification: Error bars are clearly demonstrated in both Figure 1 and Figure 2. They both have a 95\% CI

\item {\bf Experiments compute resources}
    \item[] Question: For each experiment, does the paper provide sufficient information on the computer resources (type of compute workers, memory, time of execution) needed to reproduce the experiments?
    \item[] Answer: \answerYes{} 
    \item[] Justification: Yes, we mentioned our usage of PennyLane. During the code submission, we will provide details about our code packages and how to setup. Additionally we discuss the computation resource estimate for running this on a real quantum computer.
    
\item {\bf Code of ethics}
    \item[] Question: Does the research conducted in the paper conform, in every respect, with the NeurIPS Code of Ethics \url{https://neurips.cc/public/EthicsGuidelines}?
    \item[] Answer: \answerYes{} 
    \item[] Justification: We followed all protocol listed in the NeurIPS code of Ethics

\item {\bf Broader impacts}
    \item[] Question: Does the paper discuss both potential positive societal impacts and negative societal impacts of the work performed?
    \item[] Answer: \answerYes{} 
    \item[] Justification: While Q-Policy promises significant advancements in complex decision-making tasks with potential societal benefits, its reliance on quantum phenomena also raises concerns about transparency, unpredictability, and ethical use in high-stakes applications. This idea was discussed about in the conclusion of the paper.
    
\item {\bf Safeguards}
    \item[] Question: Does the paper describe safeguards that have been put in place for responsible release of data or models that have a high risk for misuse (e.g., pretrained language models, image generators, or scraped datasets)?
    \item[] Answer: \answerNA{} 
    \item[] Justification: This paper poses no risk

\item {\bf Licenses for existing assets}
    \item[] Question: Are the creators or original owners of assets (e.g., code, data, models), used in the paper, properly credited and are the license and terms of use explicitly mentioned and properly respected?
    \item[] Answer:\answerYes{} 
    \item[] Justification: This paper used PennyLane Python Library to aid in the code. 
    Asset: PennyLane  
Version: 0.34.0  
URL: https://pennylane.ai  
License: Apache License 2.0  
Citation: Ville Bergholm, Josh Izaac, Maria Schuld, Christian Gogolin, et al. 
"PennyLane: Automatic differentiation of hybrid quantum programs." arXiv:1811.04968 [quant-ph] (2018). https://arxiv.org/abs/1811.04968


\item {\bf New assets}
    \item[] Question: Are new assets introduced in the paper well documented and is the documentation provided alongside the assets?
    \item[] Answer: \answerYes{} 
    \item[] Justification: The details of the code involved for our simulations will be available publicly on GitHub, and in supplementary materials, anonymized.

\item {\bf Crowdsourcing and research with human subjects}
    \item[] Question: For crowdsourcing experiments and research with human subjects, does the paper include the full text of instructions given to participants and screenshots, if applicable, as well as details about compensation (if any)? 
    \item[] Answer: \answerNA{}
    \item[] Justification: This research does not involve crowdsourcing nor research with human subjects.

\item {\bf Institutional review board (IRB) approvals or equivalent for research with human subjects}
    \item[] Question: Does the paper describe potential risks incurred by study participants, whether such risks were disclosed to the subjects, and whether Institutional Review Board (IRB) approvals (or an equivalent approval/review based on the requirements of your country or institution) were obtained?
    \item[] Answer: \answerNA{} 
    \item[] Justification: This research doesn't involve crowdsourcing nor research with human subjects.

\item {\bf Declaration of LLM usage}
    \item[] Question: Does the paper describe the usage of LLMs if it is an important, original, or non-standard component of the core methods in this research? Note that if the LLM is used only for writing, editing, or formatting purposes and does not impact the core methodology, scientific rigorousness, or originality of the research, declaration is not required.
    \item[] Answer:\answerNA{} 
    \item[] Justification: The core method developed in this research does not involve LLMs as any important, original, or non-standard components. 

\end{enumerate}

\end{document}